\documentclass[runningheads]{llncs}
\usepackage{graphicx}
\usepackage{textcomp}
\usepackage{amssymb}
\usepackage[nolist]{acronym}
\usepackage{subfigure}
\usepackage{amsmath}
\usepackage{booktabs}
\usepackage{multirow}
\usepackage{pgfplots}
\usepackage{boldline}
\usepackage{tabularx}
\usepackage[ruled,vlined]{algorithm2e}
\begin{document}
\begin{acronym}
\acrodef{PGD} {Projected Gradient Descent}
\acrodef{FGSM} {Fast Gradient Sign Method}
\acrodef{NFGSM} {Noise Augmented Fast Gradient Sign Method}
\acrodef{PNIL} {Pixelwise Noise Injection Layer}
\acrodef{SPSA}{Simultaneous Perturbation Stochastic Approximation}
\acrodef{KNN}{K-Nearest-Neighbour}
\acrodef{VAE}{Variational Autoencoder}
\acrodef{EOT}{Expectation Over Transformation}
\end{acronym}
\title{Towards Rapid and Robust Adversarial Training with One-Step Attacks}
%


\author{Leo Schwinn \and
Ren\'e Raab \and
Bjoern M. Eskofier}
\authorrunning{L. Schwinn et al.}
%
\institute{Friedrich-Alexander Universit\"at Erlangen-N\"urnberg (FAU), Erlangen 91056, Germany}
%
\maketitle              
\begin{abstract}
Adversarial training is the most successful empirical method for increasing the robustness of neural networks against adversarial attacks. However, the most effective approaches, like training with \ac{PGD} are accompanied by high computational complexity. In this paper, we present two ideas that, in combination, enable adversarial training with the computationally less expensive \ac{FGSM}. First, we add uniform noise to the initial data point of the \ac{FGSM} attack, which creates a wider variety of adversaries, thus prohibiting overfitting to one particular perturbation bound. Further, we add a learnable regularization step prior to the neural network, which we call \ac{PNIL}. Inputs propagated trough the \ac{PNIL} are resampled from a learned Gaussian distribution. The regularization induced by the \ac{PNIL} prevents the model form learning to obfuscate its gradients, a factor that hindered prior approaches from successfully applying one-step methods for adversarial training. We show that noise injection in conjunction with \ac{FGSM}-based adversarial training achieves comparable results to adversarial training with \ac{PGD} while being considerably faster. Moreover, we outperform \ac{PGD}-based adversarial training by combining noise injection and \ac{PNIL}.

\end{abstract}
\section{Introduction}

Deep learning has led to breakthroughs in various fields, such as computer vision \cite{Alexnet}, language processing \cite{Wavenet}, and reinforcement learning \cite{ReinforcementLearning}. However, the deployment of deep learning models in real-world applications is currently limited by their vulnerability to small adversarial perturbations to their input (adversarial examples \cite{AdversarialSzegedy}). These perturbations are produced by special optimization methods, often referred to as adversarial attacks. The magnitude of the perturbations is bounded by a value $\epsilon$ for a given norm (e.g., $l_{\infty}$, $l_{2}$) \cite{AdversarialTraining}. This guarantees that a human would still perceive the perturbed input as belonging to the original class.

\begin{figure}[t]
\centering
  \centering
  \includegraphics[width=\textwidth]{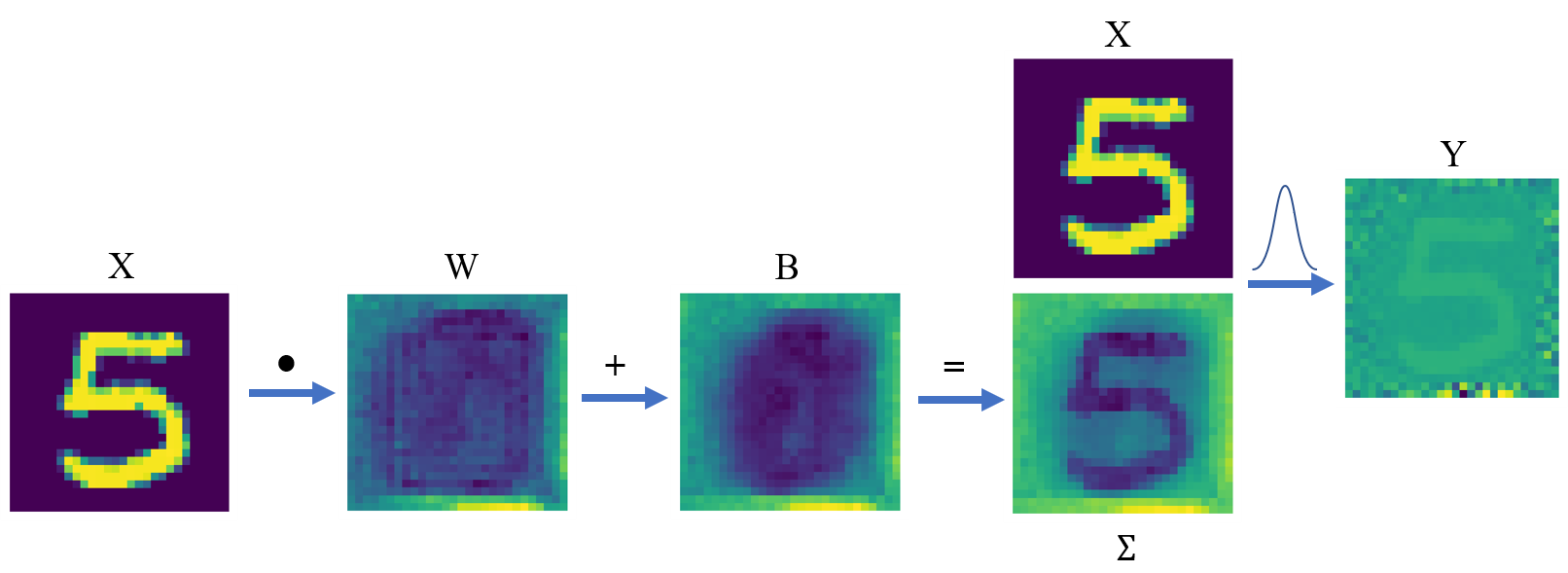}
\caption{Example of an input image $X$ from the MNIST data set, the biases $B$ and weights $W$ learned by our proposed method (\ac{PNIL}), and the resulting variance $\Sigma$ calculated for $X$. The output image $Y$ is obtained by drawing pixel values from a Gaussian distribution according to Equation \ref{Reparametrization_Eq}.}
\label{MNIST_Fig}
\end{figure}

Making neural networks robust to adversarial examples is still an unsolved problem. In a constant challenge between new adversarial attacks and defenses, most of the proposed defenses have shown to be ineffective \cite{MigatingRandomization,AdversarialRisk}.
One particularly successful approach to increase the adversarial robustness of a model is adversarial training \cite{AdversarialTraining}. This method augments the training data with adversarial examples until the model learns to classify them correctly. Here, the robustness of the model against adversarial attacks is strongly dependent on the adversarial examples used to train the model. In the past, mainly computationally expensive multi-step adversarial attacks such as \acf{PGD}, have proven effective for adversarial training \cite{TowardsAdversarialRobustness}. Less expensive one-step attacks, such as the \acf{FGSM}, have shown to be insufficient on their own \cite{EnsembleAdversarial} as they provide no adversarial robustness against stronger attacks.

In this paper, we propose a method consisting of two successive steps, that together enable fast and stable adversarial training based on the \ac{FGSM} attack.
First, we aim to find stronger perturbations with the \ac{FGSM} attack by adding noise in the range of the adversarial perturbation to the initial data point. Since the perturbation budget of the adversarial attack is limited around the noisy data point rather than the original data point, a wider variety of adversaries can be generated. In turn, the network is less likely to overfit to the particular perturbation bound.
Secondly, we improve the adversarial robustness of a neural network by including a learnable data augmentation process we call \acf{PNIL}. The \ac{PNIL} resamples the input features of a neural network from a multivariate Gaussian distribution before passing them to the next layer. The mean of this distribution is given by the input features, and the variance is calculated by the \ac{PNIL}. The resampling process weakens the adversarial perturbation, which improves the robustness of the model and increases the stability of \ac{FGSM}-based training. Furthermore, our experiments show that the additional regularization induced by the noise prevents the network from learning to obfuscate its gradients during \ac{FGSM}-based training. The \ac{PNIL} is illustrated in Figure \ref{MNIST_Fig}. We utilize the reparameterization trick, known from the Variational Autoencoder \cite{VAE}, to update the parameters of the \ac{PNIL} through backpropagation along with the other parameters of the neural network.

The contributions of this paper are the following:
\begin{itemize}
    \item We show that former claims that \ac{FGSM}-based training can lead to robust neural networks are misleading \cite{FastIsBetter}. Our experiments demonstrate that networks trained with the \ac{FGSM}-based attacks often learn to disguise their gradients, resulting in strong robustness against gradient-based attacks, but no robustness against gradient-free attacks.
    \item We propose a solution to this problem in the form of learnable data augmentation called \acf{PNIL}, which prevents gradient obfuscation of \ac{FGSM}-based training and increases the adversarial robustness of neural networks.
\end{itemize}

\section{Preliminaries}

In this section, we discuss earlier work in this research area and briefly describe common threat models in the adversarial setting.

\subsection{Related Work}

With the \acf{FGSM} \cite{AdversarialTraining}, an efficient way to construct adversarial examples with one gradient step was proposed, and used for the first form of adversarial training. In subsequent work, variants of the \ac{FGSM} have been developed, which find stronger adversaries by making multiple smaller gradient descent steps \cite{BIM,TowardsAdversarialRobustness}. Networks trained with \ac{FGSM} adversarial training proved to be vulnerable to those attacks, which made \ac{FGSM}-based training insufficient, whereas the multi-step \acf{PGD} method \cite{TowardsAdversarialRobustness} showed to be an effective way of training robust neural networks against multi-step attacks. However, the computational complexity of \ac{PGD} training is considerably larger, depending on the number of steps used to find the optimal perturbation. 

To circumvent the increased computational complexity of \ac{PGD} training, recent work tried to make \ac{FGSM}-based adversarial training feasible. The authors of \cite{EnsembleAdversarial} showed that curvature artifacts around the original data point can mask the true gradient direction, which prevents the \ac{FGSM} attack from finding a useful perturbation. Their modification of the FGSM called R+FGSM partially solved this problem by first taking a small random step with size $\alpha < \epsilon$ before performing the actual attack. Yet, their method only provided little robustness against multi-step attacks.

This idea was improved by \cite{FastIsBetter}, where the authors chose a random starting point within the full perturbation budget. This enabled adversarial training with \ac{FGSM} to achieve competitive adversarial robustness compared to \ac{PGD}-based training. Nevertheless, this approach only works with specific step sizes of the \ac{FGSM} attack and is sensitive to the network architecture. Further, it can lead to a sudden drop in model robustness and therefore requires an early stopping routine that monitors the accuracy against \ac{PGD} attacks \cite{FastIsBetter}. In our experiment, we also observed that this approach often leads to models learning to obfuscate their gradients during training, which results in no adversarial robustness against gradient-free attacks. We will call this method RFGSM in the following. 

Another approach to accelerate adversarial training is to update the parameters of the model and the adversarial perturbation simultaneously in every backward pass \cite{freeAdversarialTraining}. This combines the gradual adjustment of the perturbation and frequent updates of the model parameters. However, it has been shown that this approach does not work well with training optimization methods (e.g., cyclical learning rate \cite{CyclicLearning}, mixed precision training \cite{MixedPrecision}) and is overall slower than the RFGSM approach \cite{FastIsBetter}.

In addition to adversarial training, a variety of other defense measures have been proposed. An important branch of these methods uses random perturbations to weaken the effect of adversarial attacks \cite{RandomizedAdversarialTraining,ParametricNoiseInjection}. Most of them are based on noise injection, where noise from an isotropic Gaussian distribution is added to each feature in a layer. The parameters of the distribution were either tuneable hyperparameters \cite{RandomizedEnsemble,RandomizedAdversarialTraining} or learned through backpropagation \cite{ParametricNoiseInjection}. Intuitively, the presence of noise at inference time makes it more complicated to design an adversarial example, as the attacker does not know the specific noise when generating the adversarial perturbation. This intuition is supported by theoretical work \cite{NoiseAugmentationSmooth,NoiseTheoretical}. It was shown that injecting noise during the training phase of a neural network reduces the sensitivity of the network to small input perturbations \cite{NoiseAugmentationSmooth}. Furthermore, later work \cite{NoiseTheoretical} showed that noise injection from an exponential family gives a lower bound on the adversarial robustness of neural networks, dependent on the magnitude of the noise.  

\subsection{Adversarial Attacks}

Constructing an adversarial perturbation $r$ can be described by the following optimization problem: 

\begin{align} \label{Adversarial_Eq}
\underset{\left\Vert r \right\Vert_{p} \leq \epsilon}{\max} \mathcal{L}(F_{\theta}(X + r), Y)
\end{align}

where $X$ and $Y$ are pairs of samples in a classification task, $\mathcal{L}$ is a suitable loss function (e.g., categorical cross-entropy), and $F_{\theta}$ is a neural network parameterized by $\theta$. The manner in which the perturbation $r$ is found depends on the specific type of adversarial attack, where attacks can be characterized by the following attributes: 
\\ \\
\noindent\textbf{Norm-based:} The magnitude of the adversarial perturbation is usually bounded for a given norm ($L_{\infty}$, $L_{1}$, $L_{2}$) in such a way that any perturbed input would still be perceived as belonging to its original class by humans.
\\ \\
\noindent\textbf{Targeted/Untargeted:} Adversarial attacks are additionally grouped by their goal: They can either aim at causing any kind of misclassification (untargeted) or at causing the model to predict a particular target class (targeted).
\\ \\
\noindent\textbf{Gradient-Based/Gradient-Free:} Gradient-based attacks utilize the gradient information of the model to find the optimal perturbation (\ac{FGSM} \cite{AdversarialTraining}, \ac{PGD} \cite{TowardsAdversarialRobustness}). 
These attacks are not effective if the gradient information of the model is obscured (e.g, through a non-differentiable layer in the network). In those cases, a gradient-free method (\ac{SPSA} \cite{SPSA}) can evaluate the actual robustness of the model.
\\ \\
\noindent\textbf{White-box/Black-box:} The attack either utilizes the gradient information off the model itself to create the perturbation (white-box) or utilizes a similar model to find a perturbation which is then transferred to the original model (black-box). Black-box attacks can be used if the model is expected to hide its gradients but are generally weaker than white-box attacks.
\\ \\
\noindent\textbf{Randomized Models}:
\acf{EOT} \cite{ExpectationOverTransformation} is a proven technique for calculating gradients of models with randomized elements. It can be used to calculate gradients with respect to the expectation of the randomized component. 
\\ \\ 
\noindent Best practices for evaluating the adversarial robustness of models have been established by prior work \cite{OnEvaluatingRobustness}. One important practice is to evaluate whether models are obfuscating the gradients. If a model obfuscates its gradient, the typically applied gradient-based attacks fail. These models are however, still vulnerable to gradient-free attacks. Thus, it is important to evaluate with both gradient-free and gradient-based attacks, to get a robustness estimate for the highest risk \cite{AdversarialRisk}. Even though black-box attacks can be used to test models against gradient obfuscation, gradient-free attacks are recommended, as the transferability of black-box attacks can not be guaranteed. We will follow these practices in this paper and describe the exact threat model in Section \ref{eval}.

\section{Proposed Method}

In this section, we describe our proposed method that enables fast and stable adversarial training based on the FGSM attack.

\subsection{Noise Augmented Adversarial Training}

\begin{figure}[ht]
\centering
  \centering
  \subfigure[FGSM]{\includegraphics[width=2.0cm]{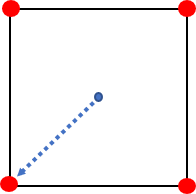}\label{FGSM_Standart_Fig}}
  \hfill
  \subfigure[R+FGSM]{\includegraphics[width=2.005cm]{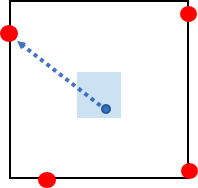}\label{FGSM_R+_Fig}}
  \hfill
  \subfigure[RFGSM]{\includegraphics[width=2.51cm]{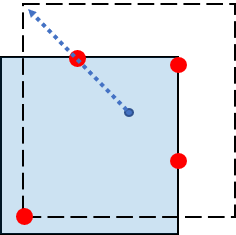}\label{FGSM_Random_Fig}}
  \hfill
  \subfigure[NFGSM]{\includegraphics[width=2.4cm]{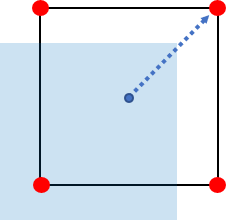}\label{FGSM_Noise_Fig}}
  \hfill
\caption{Simplified illustration of the difference between FGSM, R+FGSM, RFGSM, and the proposed \ac{NFGSM}. The potential starting point of the attack (blue dot) is given by the light blue region. An arbitrary gradient direction for each \ac{FGSM} attack is shown by a dotted arrow and the possible adversarial perturbations by a red dot at the adversarial perturbation limit (solid black line). For the RFGSM attack the step size ($1.2 \cdot \epsilon$) is additionally shown by a dashed black line. The \ac{NFGSM} attack enables more possible perturbations, by utilizing the full area given by the maximum perturbation rather than only a part of the area.}
\label{FGSM_Fig}
\end{figure}

Current research shows that adversarial robustness against strong \ac{PGD} attacks can be achieved through RFGSM-based training with only small adjustments to the initial \ac{FGSM} formulation \cite{FastIsBetter}. We argue that the limitation of the \ac{FGSM} attacks comes from the limited amount of perturbations the attack can generate. By only using the sign of the gradient, the perturbations of the standard \ac{FGSM} formulation are limited to the corners of the n-dimensional hypercube defined by the perturbation limit. The R+FGSM approach presented by \cite{EnsembleAdversarial} increases the amount of possible perturbations but still constricts perturbations to lie near the corners of the hypercube. Finally, the \ac{FGSM} variant presented by \cite{FastIsBetter} increases the amount of possible perturbations by a considerable amount. We follow the idea of increasing the variety of possible perturbations and present a similar approach, wherein we augment the training data with uniform noise before applying the \ac{FGSM} attack. We call this approach \ac{NFGSM} and give its details in Algorithm \ref{NFGSM_Algorithm}. In comparison to the RFGSM method, \ac{NFGSM} can create a wider variety of adversaries. This is achieved by moving the hypercube away from the data point. Thereby, the whole volume of the hypercube is used to create adversarial examples. In turn, the network is less likely to overfit to the particular perturbation bound used to train the model, which has been observed for prior models \cite{AdversarialMultiplePerturbations}. The difference between vanilla \ac{FGSM}, R+FGSM, RFGSM, and our method is summarized in Figure \ref{FGSM_Fig}. Keep in mind that \ac{NFGSM} is only used for training as it produces perturbations that are outside the perturbation bound of the \ac{FGSM} attack and thus could change the true label of the class for the human perception.

\begin{algorithm} \label{NFGSM_Algorithm}
\SetAlgoLined
 \For{e in [1, E]}{
   \For{$(X_{b}, Y_{b})$ in B}{
   noise = Uniform(-$\epsilon$, $\epsilon$)\;
   $X_{bn} = X_{b}$ + noise\;
   $r = \epsilon \cdot sign(\sum_{l=1}^{L} \nabla_{X_{bn}}\mathcal{L}(F_{\theta}(X_{bn}), Y_{b}))$\;
   gradient = $\nabla_{\theta}\mathcal{L}(F_{\theta}(X_{bn} + r), Y_{b})$\;
   // Update model weights $\theta$ with the given optimizer and \\
   // calculated gradient
   }
 }
 \caption{\ac{NFGSM}-based adversarial training with perturbation bound $\epsilon$, perturbation r, number of Monte Carlo simulations L number of epochs E, and batches B. An EOT based attack is given by $L>1$.}
\end{algorithm}

\subsection{Pixelwise Augmentation Layer}

To prevent the network from learning to obfuscate its gradients during the \ac{FGSM}-based training, we introduce a learnable data augmentation process we call \acf{PNIL} that can be included in any neural network. Let us consider $X_{i} \in \mathbb{R}^{n}$ as the  $i^{th}$ sample of the data set $X$ described by its features $(x_{i, 1}, x_{i, j}, ... x_{i, J})$. The \ac{PNIL} learns weights $W = (w_{1}, w_{j}, ... w_{J})$ and biases $B = (b_{1}, b_{j}, ... b_{J})$ for every input feature and uses them to calculate the variance $\Sigma_{i}$ for the respective input according to Equation \ref{PNIL_Eq}. The output of the \ac{PNIL} $X'_{i}$ is computed by sampling noise for each feature $x_{i,j}$ from a standard Gaussian distribution $\mathcal{N}(0, 1)$ which is weighted with the calculated variance $\Sigma_{i}$. The reparameterization trick enables the network to update the weights and biases of the \ac{PNIL} by backpropagation \cite{VAE}. The forward pass outlined above is given by Equation \ref{Reparametrization_Eq} and illustrated in Figure \ref{MNIST_Fig}. 

\begin{align}
\label{PNIL_Eq}
\Sigma_{i} &=  X_{i} \cdot W + B \\
\label{Reparametrization_Eq}
X'_{i} &= X_{i} + \mathcal{N}(0, 1) \cdot \exp{\frac{\Sigma_{i}}{2}}
\end{align}

Networks trained with one step attacks learn to obfuscate their gradients by converging to a global minimum where small curvature artifacts around the minimum prohibit a one-step attack from finding a strong perturbation \cite{EnsembleAdversarial}. Thereby, the networks fail to learn to defend against strong adversarial attacks but instead learn to weaken the \ac{FGSM} attack. By injecting noise to the individual input features, the \ac{PNIL} destroys the specific structure that the \ac{FGSM}-based attack generates and thus prohibits gradient obfuscation. This leads to strong robustness against gradient-free and multi-step attacks, as we show in this paper.

\begin{figure}
\centering
  \centering
  \subfigure[input]{\includegraphics[height=0.18\textwidth]{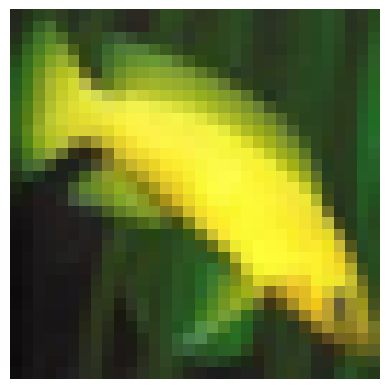}\label{SAL_Fish_Input_Fig}} 
  \subfigure[$\Sigma_{R}$ ]{\includegraphics[height=0.18\textwidth]{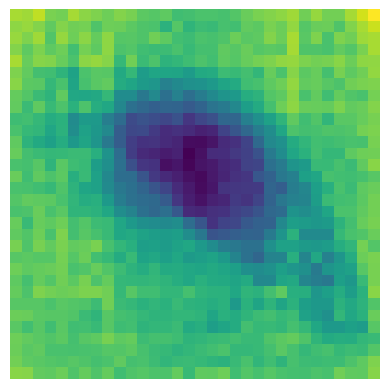}\label{PNIL_Fish_Variance1_Fig}}
  \subfigure[$\Sigma_{G}$]{\includegraphics[height=0.18\textwidth]{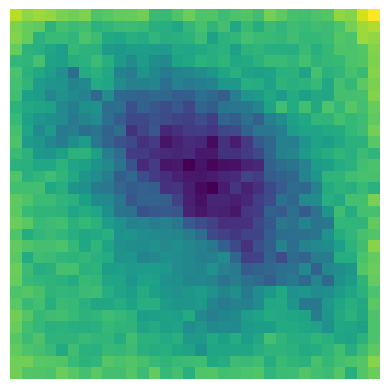}\label{PNIL_Fish_Variance2_Fig}}
  \subfigure[$\Sigma_{B}$]{\includegraphics[height=0.18\textwidth]{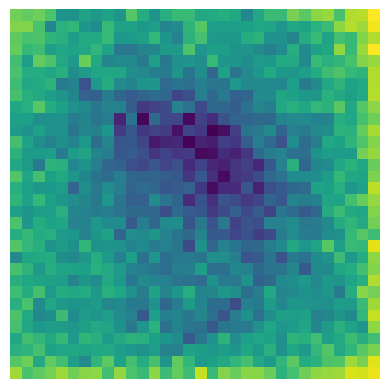}\label{PNIL_Fish_Variance3_Fig}} 
  \subfigure[output]{\includegraphics[height=0.18\textwidth]{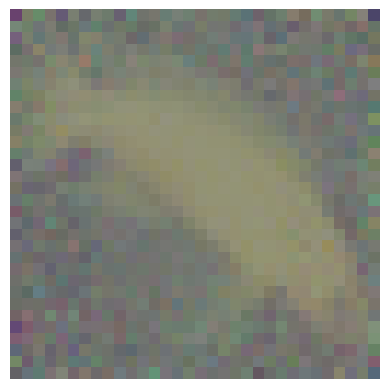}\label{PNIL_Fish_Output_Fig}} 
\caption{Example of an input image from the CIFAR10 data set, the respective variance $\Sigma$ learned by the \ac{PNIL} for all color channels, and the output image after forwarding the input trough the \ac{PNIL}. The background of the image gets lost to noise, while only the object of interest is still visible.}
\label{Cifar_Fig}
\end{figure}

Contrary to other approaches \cite{RandomizedAdversarialTraining,RandomizedEnsemble,ParametricNoiseInjection}, we learn a variance for every feature of the input with the \ac{PNIL}, instead of applying identical Gaussian noise to the features. Depending on the data set, this can lead to different values for $B$ and $W$. For simple data sets such as MNIST \cite{mnist}, the \ac{PNIL} learns low biases $B$ for the central region of the images, as the objects of interest are located there (see figure \ref{MNIST_Fig}). For more complicated data sets like CIFAR10 \cite{CIFAR}, this approach does not work, since the objects of interest are not always centered in the image. In this case, the \ac{PNIL} learns uniform biases $B$ and focuses only on the intensity of the pixels in each color channel using the weights $W$ as exemplified by Figure \ref{Cifar_Fig}. 

Note that prior work showed that architectures that apply a random transformation to the input during inference time should be evaluated with regard to the expectation of their output with the \ac{EOT} method \cite{ExpectationOverTransformation}. We will follow this approach in our evaluation.

\section{Experiments}

\subsection{Evaluation} \label{eval}

The objective of the experiments is to evaluate if noisy \ac{FGSM}-based training leads to robust neural networks and if the \ac{PNIL} further improves this robustness. Additionally, we investigate whether the \ac{PNIL} prohibits neural networks from obfuscating their gradients during the \ac{FGSM}-based adversarial training. This goal was accomplished by training pairs of nearly identical models, the only difference being that the \ac{PNIL} is added before one model as an additional first layer. All models were trained with the \ac{NFGSM} approach and with the RFGSM approach \cite{FastIsBetter} for comparison. The robustness of each model was evaluated for two different adversarial attacks (\ac{PGD}, \ac{SPSA}). The effective robustness of a neural network was then defined as the lowest accuracy against any of the attacks \cite{AdversarialRisk}. Furthermore, to increase the validity of the experiments, we train each model five times following a two-fold cross-validation scheme as proposed in \cite{dietrichttest} and apply the corrected resampled t-test as introduced in \cite{correctedttest}, to evaluate if the robustness differences between using the PNIL and not using the PNIL are significant ($\alpha$ = 0.003). Every data set used in the experiments has a predefined train and test set, whereby the predefined training sets are used for the cross-validation. For validating the training, we also split the predefined test data into two equally sized folds and assign each of them to one cross-validation fold. We point out that the average accuracy in our experiments may be lower compared to prior work, as less training data is used to train the models when using the cross-validation method. In a preliminary experiment, we show that models trained with the \ac{PNIL} exhibit lower robustness against adversaries created with the \ac{EOT} method and evaluate how many steps of \ac{EOT} are needed for reliable evaluation. 
The \ac{PNIL} is released alongside this paper to encourage other researchers to evaluate our approach independently (https://github.com/SchwinnL/ML). 

\subsection{Data}

Three different data sets were used to evaluate the adversarial robustness of the different models (MNIST \cite{mnist}, Fashion-MNIST \cite{fashionmnist}, CIFAR10 \cite{CIFAR}). All three data sets consist of images and have the goal of classifying the images according to their labels. 
\\ \\
\noindent\textbf{MNIST} consists of greyscale images of handwritten digits each of size 28x28x1 (60,000 training and 10,000 test). The MNIST data set is mainly considered a toy data set since a trivial classifier such as \ac{KNN} can achieve $95\%$ accuracy on the clean data set \cite{mnist}. However, no adversarially robust neural network architecture has been proposed for the MNIST data set yet. Thus, MNIST can be considered a non-trivial example in the adversarial case \cite{AdversarialMNIST}. 
\\ \\
\noindent\textbf{Fashion-MNIST} consists of greyscale images of 10 different types of clothing, each of size 28x28x1 (60,000 training and 10,000 test). Similar arguments as for the MNIST data set can be made for Fashion-MNIST, however the Fashion-MNIST classification task is slightly more complicated, as it contains more intricate patterns. 
\\ \\
\noindent\textbf{CIFAR10} consists of color images, each of size 32x32x3, with 10 different labels (50,000 training and 10,000 test). CIFAR10 is a subset of the ImageNet data set \cite{ImageNet} and the most challenging classification task out of the three. CIFAR10 is included in the experiments, to show that the approach generalized also to a more complicated classification task. Image data from the CIFAR10 data set was normalized to lie between zero and one. 

\subsection{Adversarial Attack Parameters} \label{Adversarial_Attack_Parameter_Section}

White-box attacks lead to stronger adversarial examples and are, in consequence, applied in preference to black-box attacks. Since the transferability of black-box attacks can not be guaranteed, we test the models for gradient obfuscation with the gradient-free \ac{SPSA} attack. All attacks are untargeted and are considered successful if the input is misclassified. We use the often-used $l_{\infty}$ norm to restrict the magnitude of the adversarial attacks in all experiments. The adversarial perturbation budget was chosen to be 8/255 for MNIST, 4/255 for Fashion-MNIST, and 0.03 for CIFAR following prior experiments in the literature \cite{TowardsAdversarialRobustness}. The RFGSM-based training was done with a step size of $1.2\cdot \epsilon$ as recommended by the authors \cite{FastIsBetter}. We used 50 gradient steps for the PGD attack and ensured that we could reach every point in the $\epsilon$-ball with a step size of $2\cdot\epsilon/50$. We use 100 steps for the \ac{SPSA} attack with a sample size of 2048 based on the results of \cite{AdversarialRisk}. We evaluated the \ac{SPSA} attack on 1000 randomly selected images for each data set, due to the considerable computational overhead of the attack. All adversarial attacks were implemented using the Cleverhans library \cite{papernot2018cleverhans} and were adjusted to work with the \ac{EOT} method.

\subsection{Model and Training Parameters} \label{Parameter_Section}

We trained two different models in our experiments. The first architecture is a small CNN, consisting of two convolutions followed by a fully-connected layer suitable for the simpler MNIST and Fashion-MNIST data sets. In addition, a residual model was used for the more complicated CIFAR10 data set (ResNet11). Residual models \cite{Residual}, have been used for adversarial benchmarks in prior literature \cite{TowardsAdversarialRobustness}, and hence provide a robust baseline for comparison. A cyclical learning rate \cite{CyclicLearning} was used together with the ADAM optimizer ($\beta_{1} = 0.9$, $\beta_{2} = 0.999$) \cite{Adam} to speed up the training. The learning rate bounds were estimated by linearly increasing the learning rate of each individual network for a few epochs, as suggested in the original paper \cite{CyclicLearning}. All models were optimized till convergence, and the checkpoint with the lowest adversarial validation loss was chosen for testing. The batch size was set to 100 for all experiments.

\section{Results} \label{Results}

The results of the experiments described in Section \ref{eval} are reported below. First, a preliminary experiment estimates the effect of the \ac{EOT} method against the \ac{PNIL}. Afterward, the impact of noisy \ac{FGSM} training and the \ac{PNIL} on the adversarial robustness of neural networks is presented. 

\subsection{Evaluating EOT Parameters}

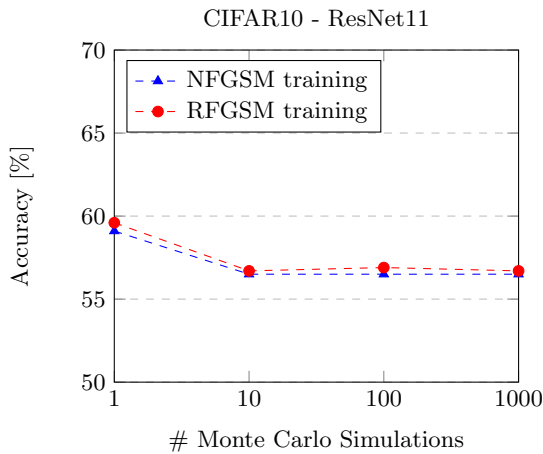
\begin{figure} 
\centering
\begin{tikzpicture}
\begin{axis}[
    height=6cm,
    title={CIFAR10 - ResNet11},
    xlabel={\# Monte Carlo Simulations},
    ylabel={Accuracy [$\%$]},
    xmin=1, xmax=4,
    ymin=50, ymax=70,
    xtick={1,2,3,4},
    xticklabels = {1, 10, 100, 1000},
    ytick={50, 55, 60, 65, 70},
    legend pos=north west,
    ymajorgrids=true,
    grid style=dashed,
]

\addplot[
    dashed,
    color=blue,
    mark=triangle*,
    mark options={solid}
    ]
    coordinates {
    (1,59.1)(2,56.5)(3,56.5)(4,56.5)
    };
    \addlegendentry{NFGSM training}
\addplot[
    dashed,
    color=red,
    mark=*,
    mark options={solid}
    ]
    coordinates {
    (1,59.6)(2,56.7)(3,56.9)(4,56.7)
    };
    \addlegendentry{RFGSM training}
\end{axis}
\end{tikzpicture}
\caption{Mean accuracy of the adversarially trained CIFAR10 models with the \ac{PNIL} against PGD adversaries calculated with different amounts of Monte Carlo simulations. The accuracy is given for \ac{NFGSM} and RFGSM-based training.}
\label{EOT_Fig}
\end{figure}

The steps necessary to perform a strong \ac{EOT}-based attack were estimated on the CIFAR10 data set against models that have a \ac{PNIL} as their first layer. The architectures and hyperparameters were chosen according to section \ref{eval}. A two-fold cross-validation was performed five times, and the mean accuracy was reported in Figure \ref{EOT_Fig}. Only a marginal decline in robustness against attacks generated with ten or more Monte Carlo simulations could be observed for all data sets. Consequently, we conclude that 100 Monte Carlo simulations are more than sufficient to get a reliable estimate of the real adversarial risk of models, which include the \ac{PNIL}.

\subsection{Adversarial Robustness}

\begin{table}[t]
\caption{Mean accuracy and standard deviation (\textpm) in percent for various configurations, with and without the \ac{PNIL} as the first layer. The perturbation strength was set to $\epsilon = 8/255$ for MNIST, $\epsilon = 4/255$ for Fashion-MNIST, and $\epsilon = 0.03$ for CIFAR10. Data from CIFAR10 are normalized between zero and one. The effective adversarial robustness of a model (lowest accuracy against any of the attacks) is shown with a bold font in the Min column. Significant differences ($\alpha = 0.003$) between the effective robustness with and without the \ac{PNIL} are highlighted by a star symbol (*). Adversarial robustness without the \ac{PNIL} is in some cases achieved through gradient obfuscation as seen by the low robustness against \ac{SPSA} attacks.}
\centering
\begin{tabularx}{\textwidth}{l|l|XXXX V{2} XXXX}
\toprule
\multirow{2}{*}{Data set} & \multirow{2}{*}{Model} & \multicolumn{4}{c V{2}}{RFGSM Training} & \multicolumn{4}{c}{NFGSM Training} \\
\cline{3-10}
& & Clean & SPSA & PGD & Min & Clean & SPSA & PGD & Min \\
\hline
\multirow{2}{*}{CIFAR10} & Basline & 75\textsubscript{\textpm 0} & 58\textsubscript{\textpm 3} & 52\textsubscript{\textpm 0} & \textbf{52}*\textsubscript{\textpm 0} & 77\textsubscript{\textpm 1} & 53\textsubscript{\textpm 3} & 50\textsubscript{\textpm 0} & \textbf{50}*\textsubscript{\textpm 0}  \\
& \multicolumn{1}{r|}{+ PNIL} & 68\textsubscript{\textpm 0} & 63\textsubscript{\textpm 1}  & 57\textsubscript{\textpm 0} & \textbf{57}*\textsubscript{\textpm 0} & 66\textsubscript{\textpm 0} & 65\textsubscript{\textpm 2} & 57\textsubscript{\textpm 0} & \textbf{57}*\textsubscript{\textpm 0} \\
\hline
\multirow{2}{*}{Fashion-MNIST} & Baseline & 83\textsubscript{\textpm 1} & 2\textsubscript{\textpm 1} & 58\textsubscript{\textpm 9} & \textbf{2}*\textsubscript{\textpm 1} & 85\textsubscript{\textpm 0} & 3\textsubscript{\textpm 2} & 64\textsubscript{\textpm 8} & \textbf{3}*\textsubscript{\textpm 2} \\
& \multicolumn{1}{r|}{+ PNIL} & 80\textsubscript{\textpm 0} & 74\textsubscript{\textpm 1} & 70\textsubscript{\textpm 0} & \textbf{70}*\textsubscript{\textpm 0} & 82\textsubscript{\textpm 0} & 78\textsubscript{\textpm 1} & 67\textsubscript{\textpm 0} & \textbf{67}*\textsubscript{\textpm 0} \\
\hline
\multirow{2}{*}{MNIST} & Baseline & 98\textsubscript{\textpm 0} & 46\textsubscript{\textpm 28} & 61\textsubscript{\textpm 25} & \textbf{46}\textsubscript{\textpm 28} & 98\textsubscript{\textpm 0} & 23\textsubscript{\textpm 36} & 54\textsubscript{\textpm 11} & \textbf{23}\textsubscript{\textpm 36} \\
& \multicolumn{1}{r|}{+ PNIL} & 98\textsubscript{\textpm 0} & 79\textsubscript{\textpm 1} & 55\textsubscript{\textpm 9} & \textbf{55}\textsubscript{\textpm 9} & 98\textsubscript{\textpm 0} & 77\textsubscript{\textpm 1} &  60\textsubscript{\textpm 4} & \textbf{60}\textsubscript{\textpm 4} \\
\bottomrule
\end{tabularx}
\label{Adversarial_Robustness_Table}
\end{table}

Initial experiments conducted on MNIST, where the \ac{PNIL} was used not only before the first layer but also multiple times throughout the network, yielded comparable results. Since using the \ac{PNIL} only at the beginning of the neural network has the lowest computational complexity. Therefore we only reported results following this approach.

The evaluation of the adversarial robustness for all configurations is given in Table \ref{Adversarial_Robustness_Table}. Within each training approach (RFGSM-based, \ac{NFGSM}-based), the effect of using the \ac{PNIL} as a first layer is analyzed. 

For both approaches, merely training with RFGSM or \ac{NFGSM} often resulted in low robustness against gradient-free attacks. Those models exhibit no real adversarial robustness and can be exploited by gradient-free attacks such as \ac{SPSA}. Networks, which included the \ac{PNIL} were not affected by gradient obfuscation, as seen by their high robustness against the \ac{SPSA} attack. Furthermore, higher effective adversarial robustness was achieved in five out of six cases when using the \ac{PNIL}, with significant differences in four cases. The two \ac{FGSM} training procedures achieved comparable results against the predefined perturbation bound.

To have a better comparison of our approach to results in the prior literature, an additional experiment was performed on the CIFAR10 data set. 
In this experiment, we compared the robustness achieved through \ac{PGD} training in \cite{TowardsAdversarialRobustness} (46\% accuracy) to our approach and achieved a mean accuracy of 59\% over ten runs. For this experiment, we used the same predefined train-test split of the CIFAR10 data set. Note that their approach used a weaker \ac{PGD} attack with 20 instead of 50 steps with the same perturbation budget $\epsilon$.

\subsubsection{Generalization:} 
\begin{figure} 
\centering
\begin{tikzpicture}
\begin{axis}[
    height=6cm,
    title={CIFAR10 - ResNet11},
    xlabel={Perturbation Strength Multiplier [$\epsilon$]},
    ylabel={Accuracy [$\%$]},
    xmin=0.5, xmax=2,
    ymin=0, ymax=100,
    xtick={0.5,0.75,1,1.25,1.5,1.75,2},
    ytick={0, 25, 50, 75, 100},
    legend pos=north west,
    ymajorgrids=true,
    grid style=dashed,
]

\addplot[
    dashed,
    color=blue,
    mark=triangle*,
    mark options={solid}
    ]
    coordinates {
   (0.50,74.3)(0.75,65.9)(1.00,50)(1.25,47.2)(1.50,41.5)(1.75,38.0)(2.00,36.1)
    };
    \addlegendentry{NFGSM training}
\addplot[
    dashed,
    color=red,
    mark=*,
    mark options={solid}
    ]
    coordinates {
  (0.50,71.6)(0.75,60.6)(1.00,52)(1.25,43.9)(1.50,39.6)(1.75,37.0)(2.00,35.1)
    };
    \addlegendentry{RFGSM training}
\end{axis}
\end{tikzpicture}
\caption{Mean accuracy of the adversarially trained CIFAR10 models without the \ac{PNIL} against PGD adversaries with varying perturbation bounds. The accuracy is given for \ac{NFGSM} and RFGSM-based training. It can be seen that \ac{NFGSM}-based training leads to slightly higher robustness for different perturbation strengths, especially for weak attacks.}
\label{pert_Fig}
\end{figure}
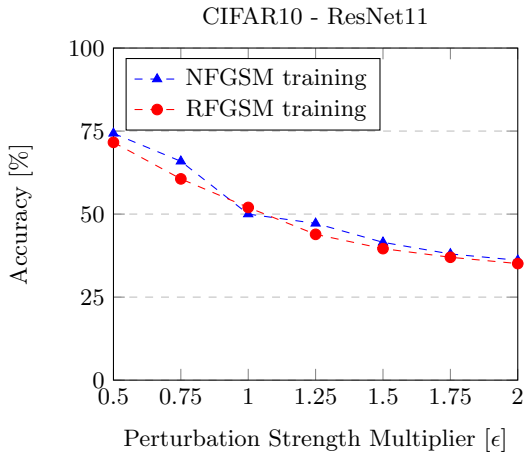

A comparison of both methods against the \ac{PGD} adversary defined in Section \ref{Adversarial_Attack_Parameter_Section} with varying perturbation bounds, shows that the \ac{NFGSM} method leads to slightly higher robustness against perturbation bounds on which the classifier was not trained on. The results are given for a baseline model trained on the CIFAR10 data set in Figure \ref{pert_Fig}.

\subsubsection{Working mechanism:} Figures \ref{MNIST_Fig} and \ref{Cifar_Fig} exemplify how the adversarial robustness is improved. Models trained with the \ac{PNIL} learn weights that lead to small variances in foreground regions and high variances in background regions. Thus, the \ac{PNIL} effectively defends against adversarial attacks that are targeting image regions, which should be irrelevant for the classification.

\section{Discussion}
Adversarial training provides a reliable method to make neural networks more robust against adversarial attacks. Unfortunately, this approach has been limited by the additional computational effort involved in training with strong adversarial attacks like \ac{PGD}. We have shown that training with \ac{FGSM} and additional input noise can lead to similar robustness. Depending on the step count of the \ac{PGD} attack, this can reduce the training time by orders of magnitude. The proposed \ac{NFGSM} approach led to a small performance improvement against adversarial attacks with varying perturbation budget compared to RFGSM-based training. Additionally, our method has the benefit that it works with a standard implementation of the \ac{FGSM} attack and is, therefore, easier to implement. Still, both methods can lead to low robustness against gradient-free attacks such as \ac{SPSA}. We proposed a solution to this problem by introducing the \ac{PNIL} and achieved significantly higher adversarial robustness by using the \ac{PNIL} as the first layer of our neural networks. We argue that the additional resampling of the input during training increases the variety of the input and makes the network less likely to obfuscate its gradients. Furthermore, during inference, the random resampling makes it harder for the attacker to construct an adversarial sample since the precise outcome of the network is not known in advance.

Contrary to variational autoencoders, no Kullback-Leibler-divergence loss is used to push the variances computed by the \ac{PNIL} in a certain direction. Intuitively, the \ac{PNIL} could learn high negative weights and biases for all features, which would lead to a zero variance in the output. As a result, the \ac{PNIL} would not change its inputs during the resampling step, and the network would be able to minimize the training loss. We argue that most pixels in an image do not contribute to the classification. Thus, only small gradients are obtained for the corresponding features in the \ac{PNIL}, and their values remain relatively unchanged. The regions where the object of interest is located contribute primarily to the classification loss and produce larger gradient values. We observe this behavior in our experiments, as seen in Figures \ref{MNIST_Fig} and \ref{Cifar_Fig}. In return, the network must focus on the crucial aspects of each image, since background information is partially lost due to noise generated by the \ac{PNIL}.

The parameters of the \ac{PNIL} in Figures \ref{MNIST_Fig} and \ref{Cifar_Fig} are determined by the fact that the objects of interest are located in the middle of the image. Consequently, the learned biases would be rendered ineffective if the relative position of an object were to change. Nevertheless, in many data sets, such as the ones used in this paper or the popular ImageNet data set \cite{ImageNet}, the objects of interest are most often found in the center of the image. To circumvent depending on the position, the variance could be calculated with another neural network. This way, the context of the pixels could be taken into account. 

Our experiments indicate that our approach enables stable \ac{FGSM}-based adversarial training, and can even outperform \ac{PGD}-based training. We assume that the increase in performance and stability will make it easier to deploy robust neural networks in industrial applications. Furthermore, we hope that researchers who rely on adversarial training will be able to use the increased performance and stability of our method to accelerate their research.

\section*{Acknowledgements}

Bjoern M. Eskofier gratefully acknowledges the support of the German Research Foundation (DFG) within the framework of the Heisenberg professorship programme (grant number ES 434/8-1).
\\ \\
\noindent The work was supported by the FAU Emerging Fields Initiative.

%
%
%
%
\bibliographystyle{splncs04}
\bibliography{bibfile}

\begin{thebibliography}{10}
\providecommand{\url}[1]{\texttt{#1}}
\providecommand{\urlprefix}{URL }
\providecommand{\doi}[1]{https://doi.org/#1}

\bibitem{RandomizedAdversarialTraining}
Araujo, A., Pinot, R., Negrevergne, B., Meunier, L., Chevaleyre, Y., Yger, F.,
  Atif, J.: Robust neural networks using randomized adversarial training. arXiv
  preprint arXiv:1903.10219  (2019)

\bibitem{ExpectationOverTransformation}
Athalye, A., Engstrom, L., Ilyas, A., Kevin, K.: {Synthesizing robust
  adversarial examples}. In: International Conference on Machine Learning,
  ICML. pp. 284--293 (2018)

\bibitem{OnEvaluatingRobustness}
Carlini, N., Athalye, A., Papernot, N., Brendel, W., Rauber, J., Tsipras, D.,
  Goodfellow, I., Madry, A., Kurakin, A.: On evaluating adversarial robustness.
  arXiv preprint arXiv:1902.06705  (2019)

\bibitem{ImageNet}
Deng, J., Dong, W., Socher, R., Li, L.J., Li, K., Fei-Fei, L.: Imagenet: A
  large-scale hierarchical image database. In: 2009 IEEE Conference on Computer
  Vision and Pattern Recognition, CVPR. pp. 248--255 (2009)

\bibitem{dietrichttest}
Dietterich, T.G.: Approximate statistical tests for comparing supervised
  classification learning algorithms. Neural computation  \textbf{10}(7),
  1895--1923 (1998)

\bibitem{AdversarialTraining}
Goodfellow, I., Shlens, J., Szegedy, C.: {Explaining and harnessing adversarial
  examples}. In: International Conference on Learning Representations, ICLR
  (2015)

\bibitem{Residual}
He, K., Zhang, X., Ren, S., Sun, J.: {Deep Residual Learning for Image
  Recognition}. In: Computer Vision and Pattern Recognition, CVPR. pp. 770--778
  (2016), \url{http://image-net.org/challenges/LSVRC/2015/}

\bibitem{ParametricNoiseInjection}
He, Z., Rakin, A.S., Fan, D.: Parametric noise injection: Trainable randomness
  to improve deep neural network robustness against adversarial attack. In:
  Computer Vision and Pattern Recognition, CVPR. pp. 588--597 (2019)

\bibitem{Adam}
Kingma, D.P., Ba, J.: Adam: {A} method for stochastic optimization. In: 3rd
  International Conference on Learning Representations, {ICLR} 2015, San Diego,
  CA, USA, May 7-9, 2015, Conference Track Proceedings (2015)

\bibitem{VAE}
Kingma, D.P., Welling, M.: {Auto-encoding variational bayes}. In: International
  Conference on Learning Representations, ICLR (2014)

\bibitem{CIFAR}
Krizhevsky, A.: Learning multiple layers of features from tiny images. Tech.
  rep. (2009)

\bibitem{Alexnet}
Krizhevsky, A., Sutskever, I., Hinton, G.E.: {ImageNet Classification with Deep
  Convolutional Neural Networks}. In: Advances in Neural Information Processing
  Systems, NeurIPS (2012)

\bibitem{EnsembleAdversarial}
Kurakin, A., Boneh, D., Tram{\`{e}}r, F., Goodfellow, I., Kurakin, A., Brain,
  G., Papernot, N., Goodfellow, I., Boneh, D., McDaniel, P.: {Ensemble
  adversarial training: Attacks and defenses}. In: International Conference on
  Learning Representations, ICLR (2018)

\bibitem{BIM}
Kurakin, A., Goodfellow, I.J., Bengio, S.: Adversarial examples in the physical
  world. In: 5th International Conference on Learning Representations, {ICLR}
  2017, Toulon, France, April 24-26, 2017, Workshop Track Proceedings (2017)

\bibitem{mnist}
LeCun, Y., Bottou, L., Bengio, Y., Haffner, P., et~al.: Gradient-based learning
  applied to document recognition. Proceedings of the IEEE  \textbf{86}(11),
  2278--2324 (1998)

\bibitem{RandomizedEnsemble}
Liu, X., Cheng, M., Zhang, H., Hsieh, C.J.: {Towards robust neural networks via
  random self-ensemble}. In: Proceedings of the European Conference on Computer
  Vision, ECCV. pp. 369--385 (2018)

\bibitem{TowardsAdversarialRobustness}
Madry, A., Makelov, A., Schmidt, L., Tsipras, D., Vladu, A.: {Towards deep
  learning models resistant to adversarial attacks}. In: International
  Conference on Learning Representations, ICLR (2018),
  \url{https://github.com/MadryLab/mnist{\_}challenge}

\bibitem{NoiseAugmentationSmooth}
Matsuoka, K.: Noise injection into inputs in back-propagation learning. IEEE
  Transactions on Systems, Man, and Cybernetics  \textbf{22}(3),  436--440
  (1992)

\bibitem{MixedPrecision}
Micikevicius, P., Narang, S., Alben, J., Diamos, G., Elsen, E., Garcia, D.,
  Ginsburg, B., Houston, M., Kuchaiev, O., Venkatesh, G., Wu, H.: Mixed
  precision training. In: International Conference on Learning Representations,
  ICLR (2018), \url{https://openreview.net/forum?id=r1gs9JgRZ}

\bibitem{ReinforcementLearning}
Mnih, V., Kavukcuoglu, K., Silver, D., Rusu, A.A., Veness, J., Bellemare, M.G.,
  Graves, A., Riedmiller, M.A., Fidjeland, A., Ostrovski, G., Petersen, S.,
  Beattie, C., Sadik, A., Antonoglou, I., King, H., Kumaran, D., Wierstra, D.,
  Legg, S., Hassabis, D.: Human-level control through deep reinforcement
  learning. Nature  \textbf{518}(7540),  529--533 (2015).
  \doi{10.1038/nature14236}

\bibitem{correctedttest}
Nadeau, C., Bengio, Y.: Inference for the generalization error. In: Advances in
  Neural Information Processing Systems, NeurIPS. pp. 239--–281 (2000)

\bibitem{Wavenet}
van~den Oord, A., Dieleman, S., Zen, H., Simonyan, K., Vinyals, O., Graves, A.,
  Kalchbrenner, N., Senior, A.W., Kavukcuoglu, K.: Wavenet: {A} generative
  model for raw audio. In: The 9th {ISCA} Speech Synthesis Workshop, Sunnyvale,
  CA, USA, 13-15 September 2016. p.~125. {ISCA} (2016)

\bibitem{papernot2018cleverhans}
Papernot, N., Faghri, F., Carlini, N., Goodfellow, I., Feinman, R., Kurakin,
  A., Xie, C., Sharma, Y., Brown, T., Roy, A., Matyasko, A., Behzadan, V.,
  Hambardzumyan, K., Zhang, Z., Juang, Y.L., Li, Z., Sheatsley, R., Garg, A.,
  Uesato, J., Gierke, W., Dong, Y., Berthelot, D., Hendricks, P., Rauber, J.,
  Long, R.: Technical report on the cleverhans v2.1.0 adversarial examples
  library. arXiv preprint arXiv:1610.00768  (2018)

\bibitem{NoiseTheoretical}
Pinot, R., Meunier, L., Araujo, A., Kashima, H., Yger, F., Gouy{-}Pailler, C.,
  Atif, J.: Theoretical evidence for adversarial robustness through
  randomization. In: Advances in Neural Information Processing Systems. pp.
  11838--11848 (2019)

\bibitem{AdversarialMNIST}
Schott, L., Rauber, J., Bethge, M., Brendel, W.: Towards the first
  adversarially robust neural network model on mnist. In: International
  Conference on Learning Representations, ICLR (2019)

\bibitem{freeAdversarialTraining}
Shafahi, A., Najibi, M., Ghiasi, M.A., Xu, Z., Dickerson, J., Studer, C.,
  Davis, L.S., Taylor, G., Goldstein, T.: Adversarial training for free! In:
  Advances in Neural Information Processing Systems, NeurIPS. pp. 3353--3364
  (2019)

\bibitem{CyclicLearning}
Smith, L.N.: Cyclical learning rates for training neural networks. 2017 IEEE
  Winter Conference on Applications of Computer Vision (WACV)  (2017).
  \doi{10.1109/wacv.2017.58}, \url{http://dx.doi.org/10.1109/WACV.2017.58}

\bibitem{SPSA}
Spall, J.C., et~al.: Multivariate stochastic approximation using a simultaneous
  perturbation gradient approximation. IEEE transactions on automatic control
  \textbf{37}(3),  332--341 (1992)

\bibitem{AdversarialSzegedy}
Szegedy, C., Zaremba, W., Sutskever, I., Bruna, J., Erhan, D., Goodfellow, I.,
  Fergus, R.: {Intriguing properties of neural networks}. In: International
  Conference on Learning Representations, ICLR (2014)

\bibitem{AdversarialMultiplePerturbations}
Tram{\`e}r, F., Boneh, D.: Adversarial training and robustness for multiple
  perturbations. In: Advances in Neural Information Processing Systems,
  NeurIPS. pp. 5858--5868 (2019)

\bibitem{AdversarialRisk}
Uesato, J., O'donoghue, B., {Van Den Oord}, A., Kohli, P.: {Adversarial Risk
  and the Dangers of Evaluating Against Weak Attacks}. In: International
  Conference on Machine Learning, ICML. pp. 5032--5041 (2018)

\bibitem{FastIsBetter}
Wong, E., Rice, L., Kolter, J.Z.: Fast is better than free: Revisiting
  adversarial training. In: International Conference on Learning
  Representations, ICLR (2020),
  \url{https://openreview.net/forum?id=BJx040EFvH}

\bibitem{fashionmnist}
Xiao, H., Rasul, K., Vollgraf, R.: Fashion-mnist: a novel image dataset for
  benchmarking machine learning algorithms (2017),
  \url{https://github.com/zalandoresearch/fashion-mnist}

\bibitem{MigatingRandomization}
Xie, C., Wang, J., Zhang, Z., Ren, Z., Yuille, A.L.: Mitigating adversarial
  effects through randomization. In: International Conference on Learning
  Representations, ICLR (2018)

\end{thebibliography}

\end{document}